# Steps Towards Programs that Manage Uncertainty*

Paul R. Cohen
Department of Computer and Information Science
University of Massachusetts
Amherst, MA 01003[†]

December 1986

## Abstract

Reasoning under uncertainty in AI has come to mean assessing the credibility of hypotheses inferred from evidence. But techniques for assessing credibility do not tell a problem solver what to do when it is uncertain. This is the focus of our current research. We have developed a medical expert system called MUM, for **M**anaging **U**ncertainty in **M**edicine, that plans diagnostic sequences of questions, tests, and treatments. This paper describes the kinds of problems that MUM was designed to solve and gives a brief description of its architecture. More recently, we have built an empty version of MUM called MU, and used it to reimplement MUM and a small diagnostic system for plant pathology. The latter part of the paper describes the features of MU that make it appropriate for building expert systems that manage uncertainty.

## 1 Introduction

This paper describes one way to give expert systems the ability that experts have to manage uncertainty, that is, to plan actions that simultaneously achieve domain goals and minimize uncertainty and its consequences. Most work on reasoning under uncertainty is concerned with assessing degrees of belief in hypotheses given evidence[1]. The primary emphasis of this paper, however, is on how expert systems should act when they are uncertain. For example: A doctor doesn't know which of two bacteremia infections a patient has, but she knows and prescribes an antibiotic effective against both. Other factors obviously play a role in the selection of antibiotics — side effects, allergic reactions, resistance, cost, and so on — but the selection is strongly constrained by the criterion that the antibiotic should cover for both disease hypotheses[2].

---

*The research was supported by DARPA-RADC Contract F30602-85-C0014 and NSF Grant IST-8409623.

[†]This paper results from ongoing research with David Day, Jeff Delisio, Mike Greenberg, Tom Gruber, Adele Howe, Rick Kjeldsen, Dan Suthers, and Dr. Paul Berman, M.D.

[1]This literature is voluminous and difficult to summarize. The reader is referred to Cohen and Gruber (1985), Szolovits and Pauker (1978), and Cohen (1985) for overviews of the field, and to past proceedings of the AAAI Workshop on Reasoning Under Uncertainty.

[2]Parallels between this example and MYCIN's therapy algorithm are intentional. MYCIN (Shortliffe, 1976) is one of the few classification problem solving systems that compensates for residual uncertainty in hypotheses by "covering" for all likely ones (Cohen, 1985).

This case illustrates four central points: first, physicians and others must make decisions and act even when uncertain; second, the precise degree of belief in hypotheses is often irrelevant to the selection of actions; third, all actions have costs — broadly construed — that must be weighed against the degree to which the action minimizes uncertainty and its consequences; and, fourth, the choice among possible actions is constrained by uncertainty. To illustrate the last point, imagine that one disease hypothesis suggests treatments A, B, and C, and the other suggests treatments B, D, and E. If the first disease hypothesis was confirmed, then a program would have to select from treatments A, B, and C. Uncertainty about the disease hypotheses relieves the program of this decision by constraining the choice of therapy to B, the only treatment that covers for both disease hypotheses.

Thus it appears that uncertainty can actually facilitate problem solving by placing constraints on actions. We can view problem-solving under uncertainty in conventional terms: uncertainty gives rise to goals that are achieved by actions. The aim of managing uncertainty is to select actions to simultaneously achieve domain goals (e.g., treat the patient) and minimize uncertainty and its consequences (e.g., be sure that the treatment will cover for several uncertain disease hypotheses). Actions can minimize uncertainty in a variety of ways: all actions produce evidence, some "hedge" over possible outcomes (e.g., the covering strategy for therapy selection), some minimize the worst-case outcome, and so on. Problem solving under uncertainty is more constrained than reasoning under the assumption of certainty because actions are selected both for their domain effects and for their effects on uncertainty[3].

We begin the paper with a detailed example of how a physician reasons with these dual criteria in medical diagnosis. Next, we briefly describe the MUM system. Finally, features of the MU architecture are discussed, and we present conclusions.

## 2 Managing uncertainty in medicine

We have explored the interplay between domain goals and the desire to minimize uncertainty in the context of medical diagnosis. We built a program (MUM) that generates *workups* for diseases that manifest as chest and abdominal pain (Cohen, Day, Delisio, Greenberg, Kjeldsen, Suthers, and Berman, 1987). A workup is a conditional sequence of questions, tests, and treatments (Fig. 1). Since the outcomes of these actions cannot be known in advance (they depend on the patient's disease), each action produces evidence. The emphasis in MUM is on getting the workup right, that is, asking only the relevant questions in the right order. MUM does not have workups like Figure 1 stored in any internal data structures; rather, its goal is to generate a sequence of questions, tests, and treatments in a way that conforms to an expert's workup. The workup graph in Figure 1 was acquired by debriefing an internist[4] for several days.

The planning that underlies a workup graph is illustrated in the context of one path through the expert workup graph, highlighted in Figure 1[5]. First, let's look at the path: The physician

---

[3]This point is reminiscent of Waltz's discovery that extending the set of line types to include shadow regions did not impede but supported line-labelling of blocks-world images (Waltz, 1975).

[4]Dr. Paul Berman, University Health Center, University of Massachusetts at Amherst.

[5]We use the word "planning" in the sense of generating an ordered sequence of actions that satisfies constraints. MUM, and more recently MU, are preliminary explorations of planning under uncertainty, where one cannot rely

asks a few questions about age, sex, and the episode of chest pain, then orders an EKG. If the EKG does not confirm angina, then the physician will prescribe therapy. If the patient's symptoms abate, an elective stress test is suggested. If the stress test suggests severe coronary artery disease an angiogram is done, possibly followed by surgery. This path contains many examples of managing uncertainty; one notable aspect is that a completely diagnostic test — the angiogram — is delayed as long as possible. Clearly, diagnostic certainty is just one of several goals that remain balanced throughout the workup.

To produce the sequence of questions highlighted in Figure 1, an expert system must reason extensively about its uncertainty and the costs and utility of evidence: The questions about age and sex cost nothing and yet potentially rule out angina. Similarly, questions about the episode of pain can rule out angina (though this is not shown in Fig. 1) and are free. The EKG can potentially confirm angina, although this is unlikely since it detects angina only if the patient is having an attack. Nonetheless, the EKG can provide other evidence (not shown in Fig. 1) pertinent to the angina hypothesis, and is very inexpensive. Thus, the first few questions are asked because they are cheap and potentially disconfirm or confirm a dangerous disease.

If the EKG is negative, the next step is to treat the patient, even though the diagnosis is not certain. Treatment satisfies several goals: it keeps the monetary cost of diagnosis low, minimizes the risk of diagnosis, protects the patient in the event his disease worsens, and provides evidence for or against the disease hypothesis. Abatement of symptoms will provide strong but not confirming evidence for angina. Despite this residual uncertainty the physician may continue treatment for months or more until the patient's condition worsens, since the treatment is inexpensive and has no serious side effects.

Treatment maintains a balance between the cost of evidence and its potential utility. Before treatment, the physician has enough evidence to give it a try. After observing relief of symptoms (and inferring the treatment to be the cause), the physician has enough evidence to continue treatment indefinitely. Treatment is the most *efficient* action at this point in the workup: other actions might produce stronger evidence (e.g., a stress test), but that evidence is stronger than necessary! If, instead of treating the patient, the physician ordered a stress test, the likelihood is that the next step would be treatment. The physician only needs enough evidence to warrant treatment, and this evidence is potentially supplied by treatment itself.

Considerable planning is required for an expert system to generate a sequence of questions, tests, and treatments that conforms to a path through a workup graph. Actions on a path must be efficient in the sense of providing adequate evidence at a given cost (in terms of money, health, quality of life, etc.) given what is already believed about a patient. This latter criterion explains why the workup delays an angiogram: a physician does not know enough about the patient's disease to perform an angiogram until the disease is practically confirmed (although the tradeoff between the cost and diagnosticity of the test, given this evidence, is acceptable.) Since the selection of actions is conditioned on evidence about disease hypotheses, one might treat a workup as a series of decisions and use conventional decision-analytic methods to select an action at each point in the workup (Raiffa, 1970; Howard, 1966). This could be tractable if the costs and utilities of actions are summarized in a single number and probability

---

on the assumption that the state of the world is known exactly, or that the effects of actions can be predicted exactly; in short, where one cannot do "dead-reckoning" planning.

374

dstributions for these numbers could be assessed. But a physician balances several criteria when planning a workup, and we believe that collapsing them into a single measure impedes progress on well-recognized AI goals such as explanation, control, and knowledge engineering (Cohen, 1985; Gruber and Cohen, 1987a). Instead, we have developed an architecture that makes control decisions by reasoning about the values of many distinct *control parameters* that affect decisions about actions is a workup graph.

## 3  The MUM System

The MUM system generates questions, tests, and treatments in a sequence that conforms to a path through an expert workup graph (Cohen, et al., 1987). Viewed structurally, an important component of MUM is a large inference net with disease-nodes at the top and data-nodes at the bottom. Intermediate *clusters,* which are clinically-significant groupings of evidence, reside in between (see Fig. 2). Data support or detract clusters, which support or detract disease hypotheses. These objects are linked by *roles of evidence*, such as **potentially-confirming** and **potentially-detracting**, that enable MUM to reason about the utility of a cluster in supporting (or detracting) the cluster or disease hypothesis above it in the inference net. Viewed functionally, MUM's task is to cycle through a loop that first establishes a focus of attention; then decides which question, test, or treatment to invoke, given the current state of the inference network; then propagates the result of this action through the inference network, updating the parameters on which control decisions depend in the next cycle. Thus, in MUM, we treat managing uncertainty as a *control* problem[6].

A closeup of part of the inference net shows that the data, clusters, and disease hypotheses have internal structure (see Fig. 3). An important component of these nodes is a local *combining function* for evidence. The level of belief in a cluster or disease hypothesis depends on the levels of belief in the clusters that support or detract from it. Local combining functions make these dependencies explicit. For example, if there is some reason to believe the patient is at risk for angina, then some support is provided for the angina hypothesis. If the EKG shows ischemic changes during an episode of substernal pain, then angina is confirmed. On the other hand, if the **postprandial** is supported, then angina is detracted. The postprandial cluster is confirmed if substernal pain occurs after eating, and it is disconfirmed if the pain is not brought on by eating.

Two aspects of the inference network support management of uncertainty through planning. First, MUM recognizes just 7 levels of belief: confirmed, strongly-supported, supported, unknown, detracted, strongly detracted, and disconfirmed. These discrete values make it easy to write local combining functions. Second, local combining functions support reasoning about the effects of gaining evidence. For example, MUM can reason that the postprandial cluster is potentially-supporting for esophageal spasm and potentially-detracting for angina. Discriminating clusters like these could be a focal point in a diagnosis because they are efficient — you get two pieces of evidence from one question. Alternatively, if the esophageal spasm hypothesis were already strongly-supported by some other evidence, MUM would reason that searching

---

[6]For related views see Erman, Hayes-Roth, Lesser, and Reddy, 1980; Clancey, 1986; Hayes-Roth, Garvey, Johnson, and Hewett, 1986.

for evidence to confirm the postprandial cluster has no marginal utility.

MUM depended on several parameters in addition to level-of-belief to select focus of attention and actions for a workup. These included the dangerousness of hypotheses and costs of tests. But although we developed a simple, declarative representation for local combining functions for levels of belief, we did not put equal effort into other control parameters, so control was difficult to specify and modify (Cohen et al., 1987). This experience led us to develop an empty version of MUM, called MU, which makes it easy to specify declaratively any control parameters required to efficiently manage uncertainty.

## 4  The MU Architecture

Structurally, MU further generalizes the inference net representation of MUM. Many control parameters can be propagated by local combining functions, just as level-of-belief was propagated with MUM's network. Functionally, MU provides the expert and knowledge engineer with a language to represent the *state* of a problem, and a set of functions to query aspects of that state. With these abilities, we have been able to easily reimplement and then modify MUM's original control structure (Delisio, Greenberg, and Kjeldsen, 1986; Gruber and Cohen, 1987b).

MU makes it easy to define new control parameters in terms of others; for example, the parameter *critical* could be defined as *at-least-supported and dangerous*. The new parameter is then associated with an object in MU's network, say *disease*, and its definition is inherited by all diseases. The *critical* parameter is composed of a static parameter and a dynamic one: the value of at-least-supported changes with level of belief, but the value of dangerous is set a priori.

MU has mechanisms for answering five classes of questions about the values of control parameters. The classes are

- Questions about current state. For example, what is the monetary cost of an EKG? is esophageal-spasm triggered? is angina currently critical?

- Questions about how to change features. For example, how can I increase the level of belief in angina? how can I decrease belief in gall-bladder at low cost?

- Focusing questions. For example, what is the set of all diseases that are both triggered and dangerous? which clusters support angina and are themselves at least supported?

- Effect questions. For example, for which diseases does age affect level of belief? does sex affect level of belief in angina?

- Questions about multiple feature changes. For example, what discriminates between angina and esophageal-spasm?

Although these examples of queries are from the medical domain, the MU architecture is domain independent. We believe that any system that manages uncertainty will need to answer at least these five classes of questions about the state of a problem.

376

MU requires the knowledge engineer to specify control strategies, but it facilitates this by providing easy access to the definitions and values of control parameters and the functions that revise their values. MU does not predispose the knowledge engineer to any particular control strategy because we believe that domains have their own bodies of expertise about how to manage uncertainty through control. Others, notably Clancey (1986), Hayes-Roth et al. (1986), and Bylander and Mittal (1986) have based control on a wide variety of declarative control parameters but have taken the additional step of specifying some aspects of control structures for broad classes of tasks. We expect that, in time, we will develop a catalog of control strategies appropriate for particular kinds of uncertain situations, but for now our focus in MU is on building the tools to acquire them (Gruber and Cohen, 1987b).

## 5 Conclusion

The main points of this paper are these:

- By emphasizing what to do when uncertain, we can view reasoning under uncertainty as a planning problem. The preconditions of actions in planning under uncertainty include states of belief, and the actions provide evidence – often by observation of their conventional effects on the state of the world. Thus actions both require and produce evidence.

- Complete certainty is not necessarily a precondition for action; for example, treatment for angina requires only a strong suspicion of the disease hypothesis. A hypothesis that is certain enough for one action, say prescribe vasodilators, may not be credible enough for a more costly action, such as coronary angiogram. In some problems, such as submarine tracking, one has almost no credible evidence and no certainty in one's model of the world. One acts nonetheless.

- The methods used to assess the degree of belief in hypotheses should be local to the hypotheses and use a few discrete levels of belief. The biggest argument for local combining functions as opposed to global ones is that if you change that latter, you are effectively redesigning your system from scratch, while local ones are easily updated (Gruber and Cohen, 1987a).

- An action typically facilitates or retards several goals, among them changes in the states of belief in one or more hypotheses. Managing uncertainty means selecting actions that balance these goals. Ths uncertainty facilitates the planning problem by providing constraints on actions.

- Managing uncertainty is not "added on" to a categorical inference system in this view. It is an integral part of the problem solving process.

The transition from MUM to MU reflects an increasing understanding of managing uncertainty through control. We will be implementing several systems in MU in the next few months to test its generality. In conclusion, we have constructed the first (MUM) and second (MU) versions of architectures for *efficient* problem-solving under uncertainty, that is, for producing solutions that maximize certainty for a given cost.

377

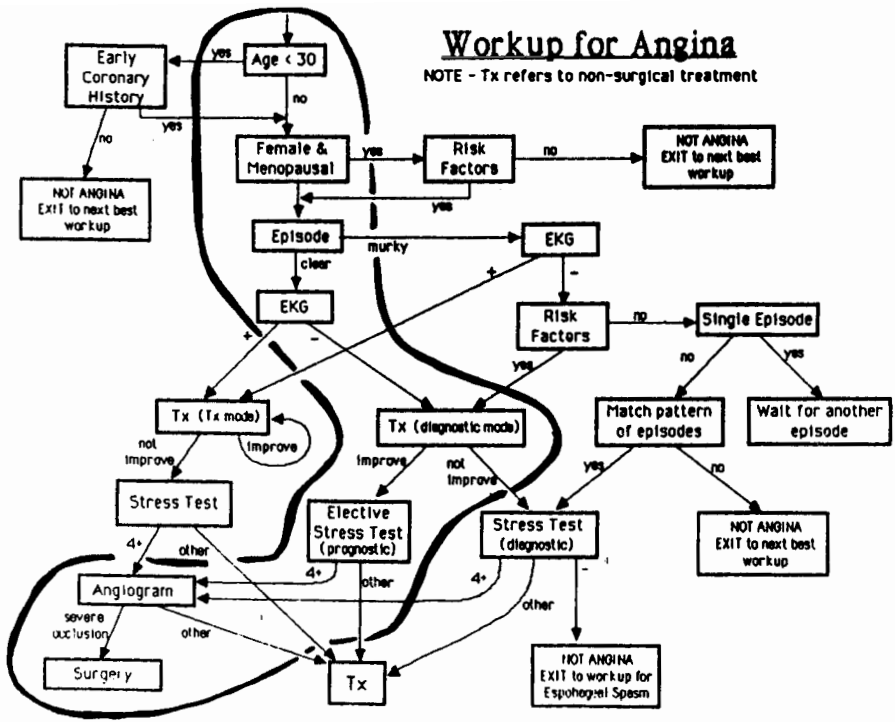

Figure 1

MUM Architecture

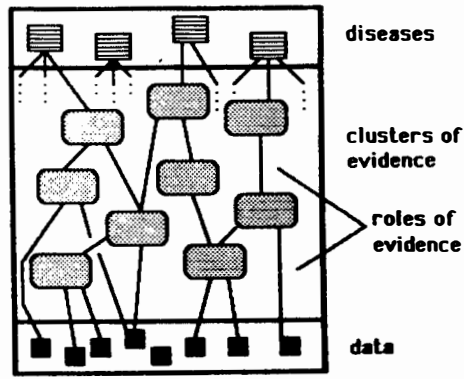

Figure 2

The problem: which to ask next given the current state of the inference network

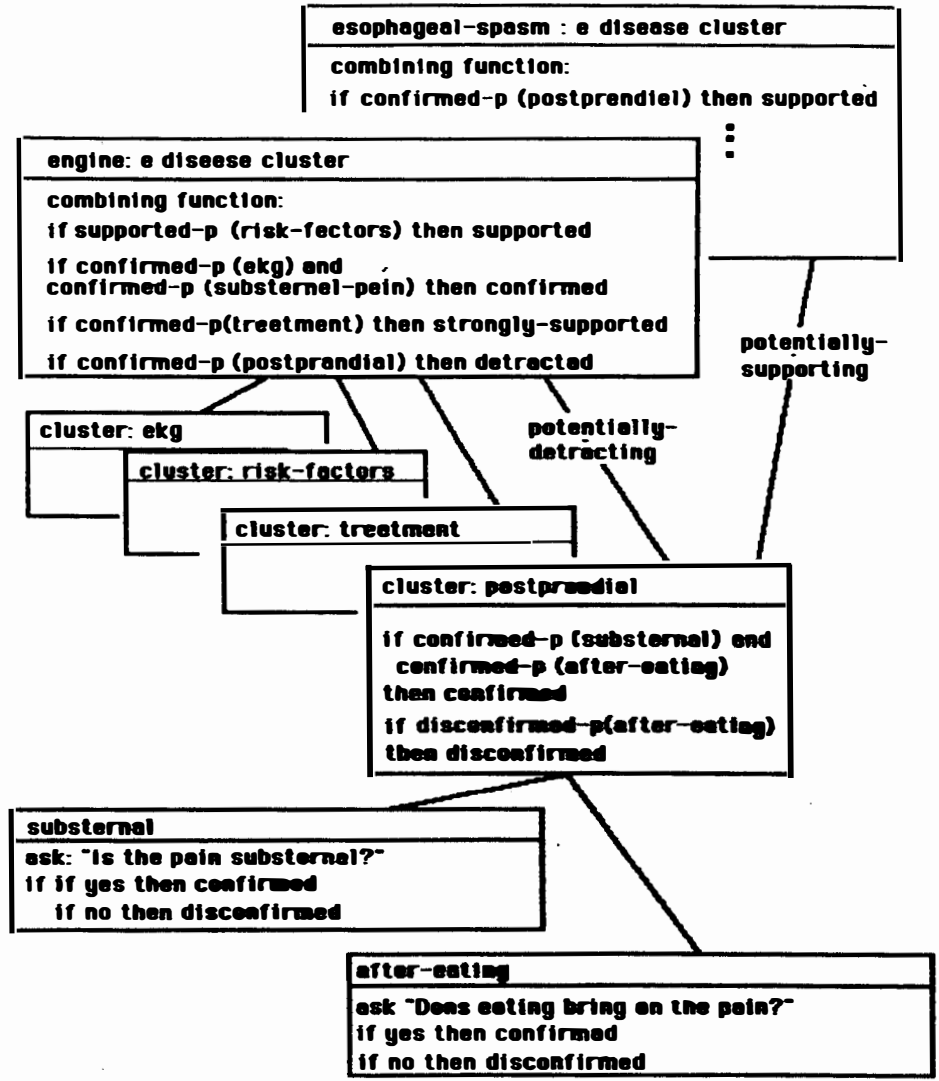

Figure 3